%% file: main.tex
\documentclass[journal]{IEEEtran}

\usepackage{mathrsfs}
\usepackage{algorithm}
\usepackage{algorithmic}
\usepackage{amsfonts}
\usepackage{amsmath}
\usepackage{cite}
\usepackage{color}
\usepackage{multirow}
\usepackage{stmaryrd}
\usepackage{dsfont}
\usepackage{graphicx} 
\usepackage{multirow}
\usepackage{threeparttable}

\usepackage{booktabs}
\usepackage{adjustbox}

\usepackage{longtable}

\usepackage{graphicx,times,amsmath} 

\begin{document}
\title{A Survey on Facial Image Privacy Preservation in Cloud-Based Services}

\author{Chen Chen, Mengyuan Sun, Xueluan Gong,~\IEEEmembership{Member,~IEEE,} Yanjiao Chen,~\IEEEmembership{Senior Member,~IEEE,} and Qian Wang,~\IEEEmembership{Fellow,~IEEE} 

\thanks{C. Chen and X. Gong are with Nanyang Technological University, Singapore. E-mail: \{chen.chen, xueluan.gong\}@ntu.edu.sg.}


\thanks{M. Sun and Q. Wang are with the School of Cyber Science and Engineering, Wuhan University, China.  E-mail: \{mengyuansun, qianwang\}@whu.edu.cn.}

\thanks{Y. Chen is with the College of Electrical Engineering, Zhejiang University, China. Email: chenyanjiao@zju.edu.cn.}

%
}

\maketitle

\input{abstract}

\input{intro}

\input{preliminaries}

\input{evaluations}

\input{comparison}

\input{future}

\input{conclusion}

\bibliographystyle{IEEEtran}
{\color{black}\bibliography{sigproc}}

\begin{IEEEbiographynophoto}
{Chen Chen} received his B.S. degree from the University of Science and technology Beijing, China, in 2012. He received his Master of Computer Science from the University of New South Wales, Australia, in 2018 and his Ph.D. degree in Computer Science from Nanyang Technological University, Singapore, in 2024. His research interests lie in the area of Network Security, AI Safety, Knowledge Graphs and Large Language Models.
\end{IEEEbiographynophoto}

\begin{IEEEbiographynophoto}{Mengyuan Sun} is currently pursuing her bachelor degree in the School of Cyber Science and Engineering at Wuhan University, China. Her research interests include network security and AI security.
\end{IEEEbiographynophoto}


\begin{IEEEbiographynophoto}
{Xueluan Gong} received her B.S. degree in Computer Science and Electronic Engineering from Hunan University in 2018. She received her Ph.D. degree in Computer Science from Wuhan University in 2023. She is currently a Research Fellow at the School of Computer Science and Engineering at the Nanyang Technological University, Singapore.
Her research interests include network security and AI security. She has published more than 30 publications in top-tier international journals or conferences, including IEEE S\&P, NDSS, ACM CCS, Usenix Security, WWW, ACM Ubicomp, IEEE JSAC, TDSC, TIFS, etc. 
\end{IEEEbiographynophoto}

\begin{IEEEbiographynophoto}
{Yanjiao Chen} received her B.E. degree in Electronic Engineering from Tsinghua University in 2010 and Ph.D. degree in Computer Science and Engineering from Hong Kong University of Science and Technology in 2015. She is currently a Bairen researcher in Zhejiang University, China. Her research interests include spectrum management for Femtocell networks, network economics, network security, AI security, and Quality of Experience (QoE) of multimedia delivery/distribution.
\end{IEEEbiographynophoto}

\begin{IEEEbiographynophoto}{Qian Wang} is a Professor in the School of Cyber Science and Engineering at Wuhan University, China. He was selected into the National High-level Young Talents Program of China, and listed among the World's Top 2\% Scientists by Stanford University. He also received the National Science Fund for Excellent Young Scholars of China in 2018. He has long been engaged in the research of cyberspace security, with focus on AI security, data outsourcing security and privacy, wireless systems security, and applied cryptography. He was a recipient of the 2018 IEEE TCSC Award for Excellence in Scalable Computing (early career researcher) and the 2016 IEEE ComSoc Asia-Pacific Outstanding Young Researcher Award. He has published 200+ papers, with 120+ publications in top-tier international conferences, including USENIX NSDI, IEEE S\&P, ACM CCS, USENIX Security, NDSS, ACM MobiCom, ICML, etc., with 20000+ Google Scholar citations. He is also a co-recipient of 8 Best Paper and Best Student Paper Awards from prestigious conferences, including ICDCS, IEEE ICNP, etc. In 2021, his PhD student was selected under Huawei's  ``Top Minds'' Recruitment Program. He serves as Associate Editors for IEEE Transactions on Dependable and Secure Computing (TDSC) and IEEE Transactions on Information Forensics and Security (TIFS). He is a fellow of the IEEE, and a member of the ACM.
\end{IEEEbiographynophoto}



\end{document}

%% file: abstract.tex
\begin{abstract}

Facial recognition models are increasingly employed by commercial enterprises, government agencies, and cloud service providers for identity verification, consumer services, and surveillance. These models are often trained using vast amounts of facial data processed and stored in cloud-based platforms, raising significant privacy concerns. Users’ facial images may be exploited without their consent, leading to potential data breaches and misuse. This survey presents a comprehensive review of current methods aimed at preserving facial image privacy in cloud-based services. We categorize these methods into two primary approaches: image obfuscation-based protection and adversarial perturbation-based protection. We provide an in-depth analysis of both categories, offering qualitative and quantitative comparisons of their effectiveness. Additionally, we highlight unresolved challenges and propose future research directions to improve privacy preservation in cloud computing environments.

\end{abstract}
\begin{IEEEkeywords}
Facial image privacy, cloud-based services,  image obfuscation, adversarial perturbation, and facial privacy protection.
\end{IEEEkeywords}

%% file: intro.tex
\section{Introduction}

The rapid growth of facial recognition technology has greatly improved convenience in applications like payment systems, mobile device unlocking, and access control \cite{shivanna2024biometric}. Cloud-based services, which store and process large amounts of user data, increasingly integrate facial recognition to streamline these functions. However, as users depend more on cloud platforms, their facial images face heightened risks of unauthorized access, misuse, and exploitation. Cloud service providers often retain facial images for purposes such as identity verification, model training, and user authentication, raising serious concerns about data privacy and security \cite{zhang2024validating}.

The extensive collection of facial images in cloud environments facilitates identity recognition but also introduces significant privacy risks. Unauthorized access to this sensitive data can result in deepfake creation, identity theft, and other malicious activities. These risks are further amplified in cloud platforms, where data is frequently shared across multiple entities and regions, making it harder for users to control how their facial data is used.

\begin{figure}[tt]
	\centering
	\includegraphics[width=8.4cm]{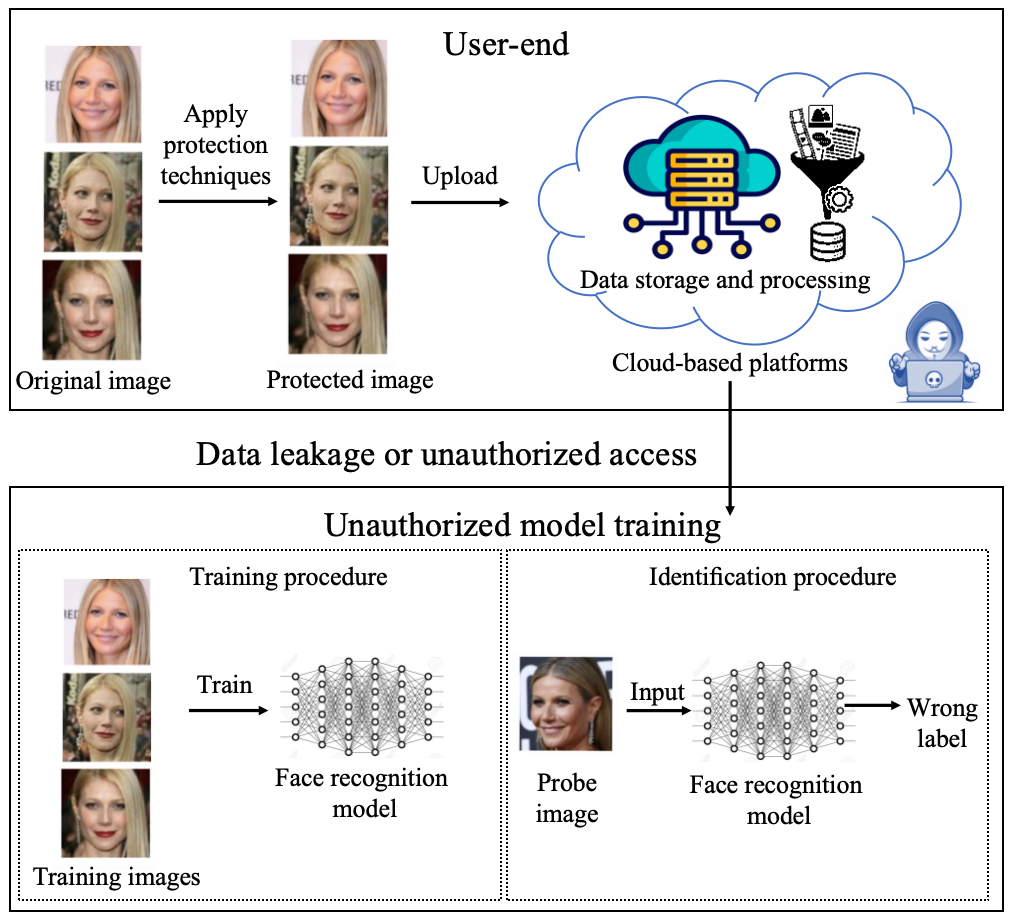}\\
	\caption{{\color{black}Protecting private facial privacy from unauthorized exploitation in cloud-based services. }}
    \label{fig:overview}
    \vspace{-0.6cm}
\end{figure}


There is a significant body of research exploring how to protect the privacy of facial images stored and processed in cloud-based services \cite{zhong2022opom, hu2022protecting, yang2021towards,li2021deepblur}. Before uploading private facial photos to cloud platforms for storage or processing, users can add perturbations to these images to make them ``unlearnable" by commercial face recognition models. Even when large volumes of such protected images are processed in the cloud, the performance of face recognition models degrades to a level similar to random guessing when attempting to recognize a new photo (probe image) of the same individual. Preserving users' facial privacy while ensuring the functionality involves addressing both privacy and utility concerns. Privacy entails making the user's identification information unrecognizable in the uploaded face images, while utility focuses on maintaining the visual and functional quality of these images. To strike this balance, users can apply imperceptible noise to their facial images, utilize makeup techniques, or blend their facial features with other publicly available images to mask identifying characteristics.


In this article, we present a comprehensive and systematic review of methods for preserving facial image privacy in cloud-based services. We categorize facial privacy protection methods into two main types: image obfuscation-based methods and adversarial perturbation-based methods, based on how the protected images are generated. We summarize the strengths and weaknesses of these techniques and provide qualitative and quantitative comparisons of various approaches. Finally, we identify key challenges and propose future research directions to further enhance privacy protection in cloud computing environments.


%% file: preliminaries.tex
\section{Preliminaries}

\subsection{Face Recognition}




Face recognition is a biometric technology that identifies individuals by analyzing unique facial features. It compares the facial characteristics in an image against a database of known faces to find a match. Typically, face recognition models extract features such as facial geometry, texture, and color, and transform them into a compact representation called a face embedding. These embeddings are then compared to embeddings in the database to recognize or verify an individual’s identity.

Several types of face recognition models are widely employed in real-world applications, including convolutional neural network (CNN)-based models, DeepFace-based models, and FaceNet-based models. Each model type offers distinct advantages in terms of accuracy, speed, and robustness. Generally, deep learning-based models such as CNNs and FaceNets have set new benchmarks for facial recognition performance and are frequently utilized in cloud-based services for tasks such as identity verification, access control, and facial authentication.

The accuracy of facial recognition systems can be affected by factors such as lighting conditions, facial expressions, and aging. To address these challenges, many cloud-based facial recognition systems integrate multiple algorithms to optimize accuracy. These systems are now being used in diverse applications, from security and surveillance to mobile device authentication and cloud-based services like photo organization. However, the widespread use of facial recognition raises significant privacy concerns, prompting regulatory bodies to impose guidelines and restrictions to protect user privacy in cloud computing environments.

\subsection{Cloud-Based Services and Facial Data Processing}
Cloud-based services provide vast storage and processing capabilities for handling sensitive user data, including facial images. These services are increasingly used for applications like identity verification, remote authentication, and user profile management. Cloud platforms offer scalable solutions, allowing enterprises to process facial data in real-time and across distributed regions, making them integral to modern biometric systems.

However, this convenience comes with significant privacy risks. Cloud platforms typically handle facial images for various purposes, including training facial recognition models and enabling user authentication. Given the sensitivity of facial data, unauthorized access to these images can lead to privacy breaches, identity theft, or other malicious activities, such as deepfake creation. Moreover, since cloud services often span multiple geographic locations, it is challenging for users to maintain control over how and where their facial data is stored and processed.

To address these concerns, researchers have proposed various privacy-preserving methods for facial images in cloud environments. These methods aim to protect user privacy without compromising the utility of the data in cloud-based applications. Two main categories of protection techniques are frequently explored: image obfuscation-based methods and adversarial perturbation-based methods. These methods ensure that facial images remain secure, even when processed by cloud-based face recognition systems.

%% file: evaluations.tex
\section{State-of-the-art Face Image Protection Approaches}

As the use of cloud-based services for storing and processing facial data continues to expand, the need for robust privacy protection techniques becomes more critical. To address privacy concerns in cloud environments, researchers have developed several state-of-the-art methods to protect facial images from unauthorized exploitation. These techniques are primarily categorized into image obfuscation-based protection methods and adversarial perturbation-based protection methods. Both approaches aim to preserve the utility of facial images while making them unrecognizable or ``unlearnable" to commercial face recognition models used in cloud services.

\subsection{Image Obfuscation-based Protection Methods}
Image obfuscation methods alter or mask specific characteristics of facial images to make them unrecognizable to facial recognition systems. These methods typically involve modifying identifiable facial features or blending them with other visual elements to prevent accurate face identification.

\subsubsection{Specific properties obfuscation}
Techniques in this category focus on obfuscating certain facial attributes, such as gender, age, or skin texture. For example, AMT-GAN \cite{hu2022protecting} uses generative adversarial networks (GANs) to apply makeup to a user’s face, making it difficult for face recognition models to accurately identify the individual. This method preserves the overall appearance of the face while making critical identity features difficult to detect. However, this technique is primarily effective for female faces, and adapting it for broader use cases remains a challenge. To address this challenge, Tang et al. \cite{tang2022gender} proposed Gender-AN, a gender-adversarial network that directly confuses gender to protect facial privacy. The face generator in Gender-AN undergoes optimization through a multi-task-based loss function, incorporating attribute manipulation loss, face matcher loss, adversarial loss, and reconstruction loss functions. This optimization approach ensures robust generalization performance while maintaining the natural appearance.

\subsubsection{Whole image obfuscation}
Li et al. \cite{li2021deepblur} proposed DeepBlur, which blurs images in the latent space of a pre-trained generative model capable of creating realistic facial images. The process involves three steps: latent representation search, deep blurring, and image generation. The input face image is transformed into a latent space vector, modified multiple times using a pre-trained model to remove identifying information, and then a realistic face is generated based on the blurred latent space vector model. This approach allows for generating different levels of obfuscation for various applications. However, there is a noticeable visual gap between the generated and original images.

He et al. \cite{he2024diff} introduced Diff-Privacy, a more advanced diffusion-based method for anonymizing facial images in cloud-based environments. Diffusion models, originally designed for image generation and restoration, incrementally add noise to an image and then reverse the process to recover or generate a new image. In Diff-Privacy, this diffusion process is employed to introduce noise into facial images in such a way that the images remain unrecognizable by facial recognition systems. Unlike DeepBlur, which directly manipulates the latent space, Diff-Privacy operates at the pixel level to iteratively transform the image. The method maintains high visual fidelity while making the facial data ``unlearnable" by machine learning models. This flexibility allows it to meet varying privacy requirements in cloud-based services, where balancing visual quality and data privacy is critical.

In addition to software-based solutions, Lopez et al. \cite{lopez2024privacy} presented a hardware-level approach to facial privacy protection through the use of optical encoders. Their method generates face heatmaps that obscure the user's identity while preserving other aspects of the image for non-sensitive applications. This hardware-level solution adds another layer of protection, offering a more robust defense against identity-based recognition systems by directly addressing the vulnerabilities that software-based solutions may overlook in cloud environments.

Zhang et al. \cite{zhangtransferable} further addressed the issue of Blind Face Restoration (BFR), where obfuscated facial images could be restored using advanced restoration models. Their transferable adversarial obfuscation method creates obfuscated images that resist restoration attempts by BFR models. This method enhances the generalization of obfuscated faces across various BFR models, ensuring that even if adversaries attempt to retrain restoration models on obfuscated images, the obfuscation remains effective. This technique is particularly valuable in scenarios where attackers have access to BFR systems and aim to reverse facial obfuscation.

\subsection{Adversarial Perturbation-based Protection Methods}

Adversarial perturbation techniques focus on adding subtle, imperceptible noise to facial images, making them resistant to recognition by machine learning models without significantly altering the visual quality of the image. These perturbations ensure that even if the images are used to train a face recognition model, the model will fail to learn meaningful representations from the data. We divide the schemes of adversarial perturbation into two categories according to whether requiring auxiliary data or not.

\subsubsection{Non-auxiliary data requirement} 

{\color{black} Under such a protection paradigm, the defenders do not rely on any auxiliary dataset to protect the private faces.}
Huang et al. \cite{huang2021unlearnable} generated unlearnable examples that contain error-minimizing noises, referred to as EM. It is based on such an intuition: the samples with larger training loss may contain more knowledge to be learned. Generating optimal perturbations that minimize the loss can suppress the informative knowledge of the data, thereby protecting the data from being learned. 
Further, Fu et al. \cite{fu2021robust} introduced REM (robust error-minimizing noise), which effectively reduces the adversarial training loss and suppresses the learnable knowledge of data in adversarial training.

{\color{black}Zhong et al. \cite{zhong2022opom} presented OPOM (one person one mask) to generate class-wise masks by optimizing each training sample in a direction away from the feature subspace of the protected face image. {\color{black}The feature subspace represents the specific set of features or representations associated with that face, often determined by techniques like class centers, convex hulls, and affine hulls.} OPOM is efficient and suitable for real-time privacy protection applications. However, OPOM may not be effective in scenarios where there are significant differences between the testing images and the training samples, such as larger poses, different illumination, or different occlusions compared to the original protected image.} 

Shamshad et al. \cite{shamshad2024makeup} introduced DFPP, which leverages makeup as an adversarial technique to obfuscate identity in facial images. DFPP employs a deep neural network to transfer makeup features from a reference image to a target image, effectively masking the user's identity while preserving visual aesthetics. This method is particularly effective in black-box facial recognition environments, where the internal workings of the recognition model are unknown. Although makeup is the medium, the technique generates adversarial perturbations that disrupt the recognition model, making it fall under adversarial perturbation-based methods rather than specific property obfuscation.


\begin{figure}[tt]
\vspace{-0.4cm}
	\centering
	\includegraphics[width=9cm]{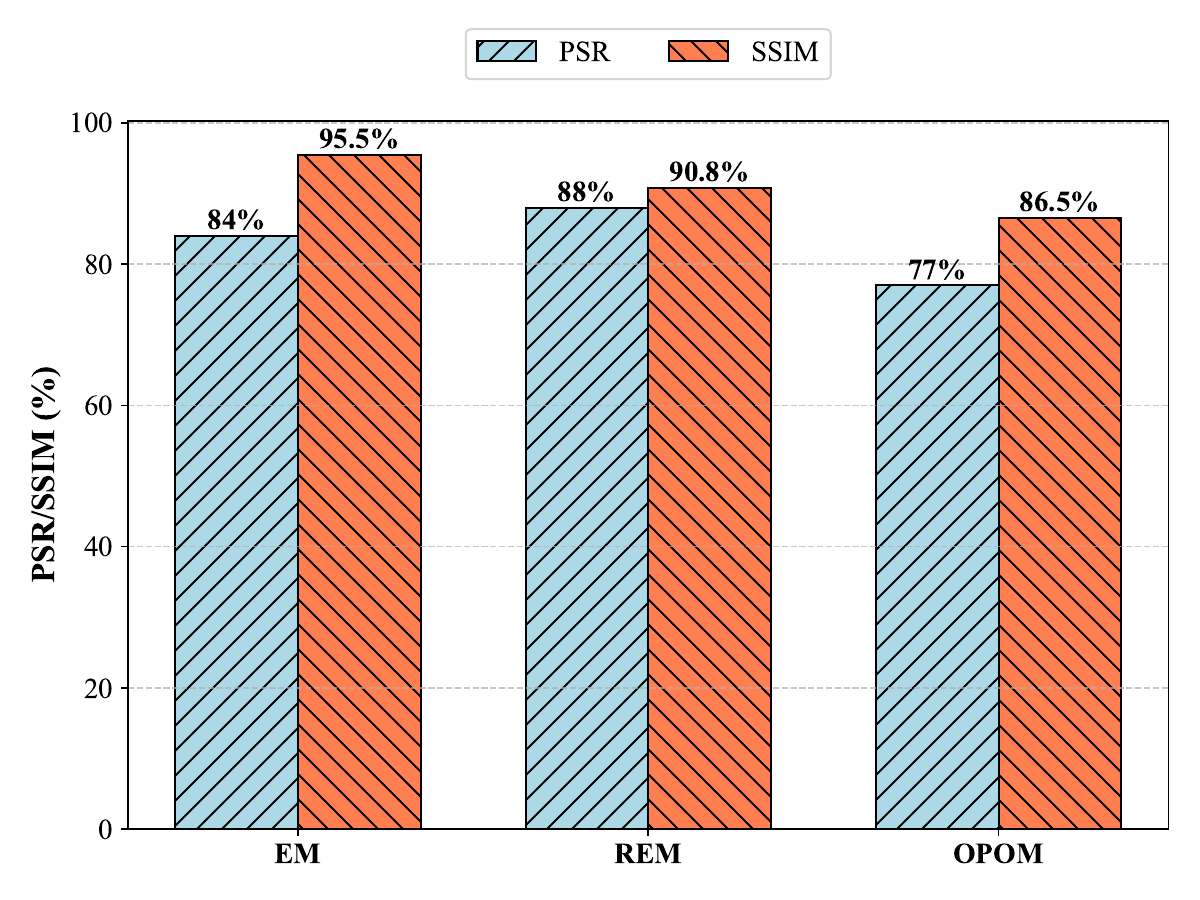}\\
	\caption{Comparison of facial privacy protection performance of EM \cite{huang2021unlearnable}, REM \cite{fu2021robust}, and OPOM \cite{zhong2022opom}.}
    \label{fig:fl}
    \vspace{-0.4cm}
\end{figure}

\begin{figure}[tt]
	\centering
	\includegraphics[width=9cm]{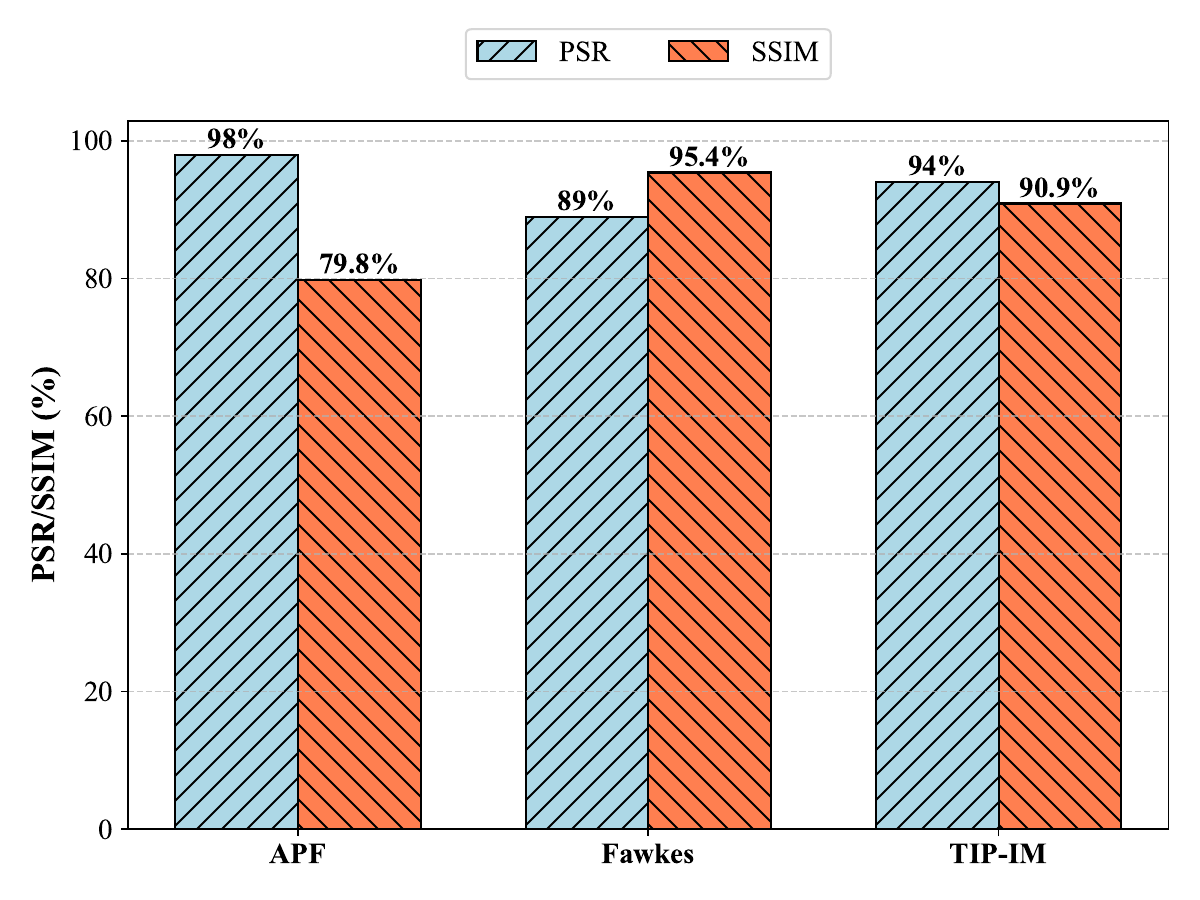}\\
	\caption{Comparison of facial privacy protection performance of APF \cite{zhang2020adversarial}, Fawkes \cite{shan2020fawkes}, and TIP-IM \cite{yang2021towards}.}
    \label{fig:2}
    \vspace{-0.6cm}
\end{figure}

\begin{figure*}[tt]
	\centering
	\includegraphics[width=13cm]{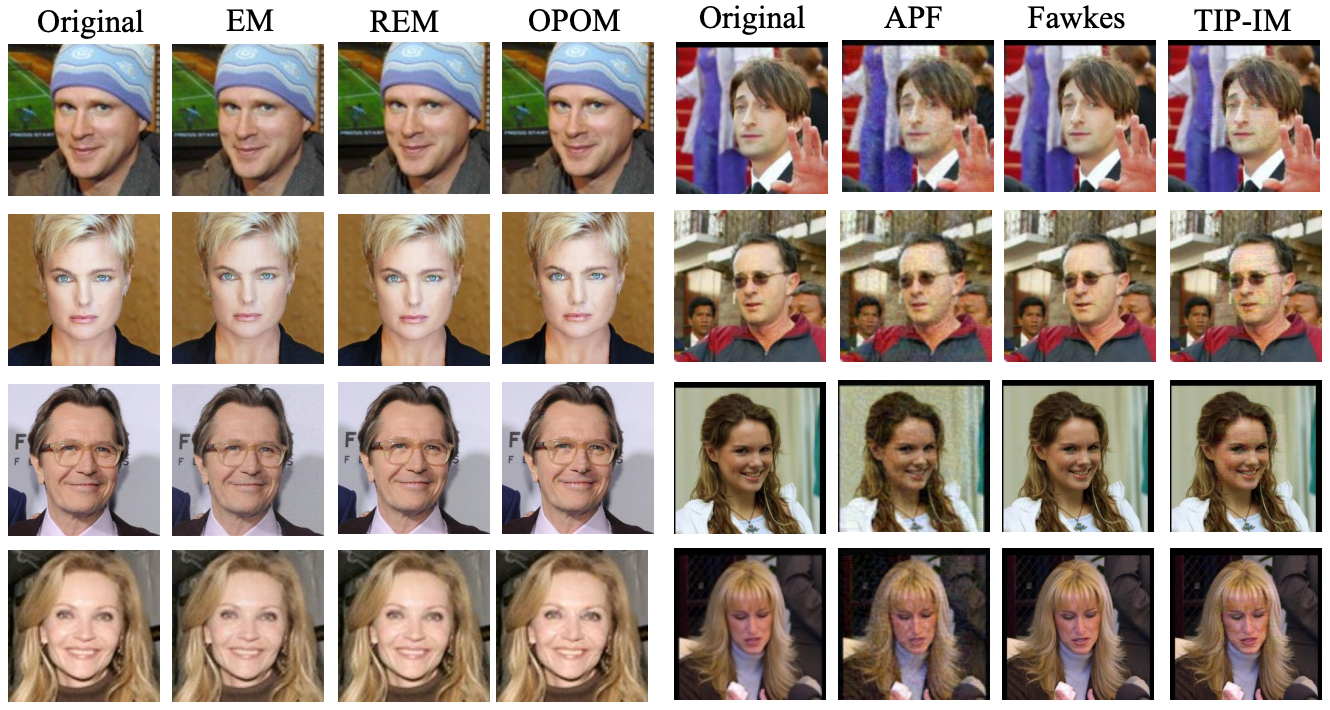}\\
	\caption{Visual comparison results of different facial privacy protection methods.}
    \label{fig:c}
    \vspace{-0.6cm}
\end{figure*}

\subsubsection{Auxiliary data requirement}

{\color{black}In this case, the defender protects private faces with the assistance of an auxiliary dataset.}


{\color{black}
Zhang et al. \cite{zhang2020adversarial} developed APF (adversarial privacy-preserving filter) to protect online shared face images from unauthorized use. APF is an end-cloud collaborated adversarial attack framework and ensures that only users' own devices can access the original images. It involves three key modules: image-specific gradient generation, adversarial gradient transfer, and universal adversarial perturbation enhancement. Image-specific gradient generation extracts gradients on the user end, and adversarial gradient transfer fine-tunes them in the server cloud. The server creates a universal perturbation by maximizing feature distances in an auxiliary dataset. An enhancement module adjusts it to match the image-specific perturbation for better integration.}
{\color{black}In the same period, Shan et al. \cite{shan2020fawkes} proposed Fawkes. In Fawkes, the defender first selects a target category from a publicly auxiliary dataset. Specifically, the authors use a feature extractor to calculate the centroid of the feature space for each candidate class and pick the class whose feature representation centroid is the most dissimilar to the feature representation of all the private images. Then Fawkes randomly selects an image from the target class and optimizes the calculation of the small disturbance of the protected image. In this step, the authors introduced DSSIM (Structural Dissimilarity Index) to ensure that the processed image is visually similar to the original image. However, Fawkes is designed for the white box, requiring access to the feature extractor the user wishes to subvert.}

{\color{black}Unlike Fawkes, Yang et al. \cite{yang2021towards} introduced TIP-IM (targeted identity-protection iterative method) to encrypt private facial images, offering privacy protection against black-box face recognition systems. TIP-IM generates adversarial identity masks and overlays them on facial images, concealing the original identities without compromising visual quality. To create the adversarial identity masks, the authors employ a novel optimization mechanism, using MMD (Max Mean Discrepancy) loss to enhance the visual quality of the crafted adversarial examples. This allows concealing the original images without sacrificing their visual quality, and the generated masks are effective for other black-box models. To control the misclassification results, the authors use greedy insertion to select the optimal victim image from an auxiliary dataset. However, as TIP-IM lacks constraints on visual behavior, except for naturalness defined by MMD, the increase in noise intensity may not guarantee a natural change in pixel information, especially when the de-identification effect dominates naturalness in the face image.}

%% file: comparison.tex
\section{Comparison and Evaluation}\label{section:comparision}

\subsection{Qualitative Comparison}
We compare the existing image obfuscation-based protection and adversarial perturbation-based protection methods in Table~\ref{tab:my-table1} and Table~\ref{tab:my-table2}, respectively. 
\begin{itemize}
    \item \emph{API}:
    Commercial facial recognition APIs are widely used by businesses, law enforcement, private entities, and governments. These APIs are typically based on large datasets and are known for their high recognition accuracy. A face privacy protection method that can successfully mislead these APIs is considered to be practical in real-world scenarios. Methods such as AMT-GAN, DeepBlur, and TIP-IM are shown to be effective in protecting against services like Aliyun Face, Microsoft Face, and the Tencent AI Open Platform.

    \item \emph{Open-set}: 
    In close-set face recognition, the faces used for testing are limited to those identities that are present in the training set, and no individuals outside of the training set will be included. However, in most real-world applications, the face to be recognized is likely to be an identity that is not present in the training set, and in such cases, open-set face recognition is necessary. We can see that only a few methods (e.g., AMT-GAN, Diff-Privacy, and Optics) are effective in open-set face recognition scenarios.


\item \emph{Real-time capability}:
This metric measures whether the obfuscation method can be applied in real-time. Optics and Diff-Privacy are both capable of real-time application, making them suitable for scenarios such as live video streams. Methods like AMT-GAN and Gender-AN require more processing time, limiting their applicability in real-time systems.

\item \emph{Scalability}:
Scalability measures the method's efficiency when applied to large-scale datasets in cloud environments. Diff-Privacy and DeepBlur show high scalability, as they can be applied efficiently across many images in cloud settings. Optics, being a hardware-based solution, is less scalable due to the need for specific equipment.

\item \emph{Hardware compatibility}:
Methods like Optics require specific hardware for implementation, which may limit their widespread use. Other methods, such as DeepBlur and Diff-Privacy, are software-based, making them more flexible and easier to deploy across various platforms.


\item \emph{Target-identity}: When creating an adversarial mask, the defender can connect it to a specific target identity for misclassification or any random misclassification identity. It is shown that only Fawkes and TIP-IM belong to target-identity scenarios, where any samples protected by the generated mask will be misclassified into a pre-designed target identity.

 \item \emph{Person-specific}: A person-specific mask is specific to an individual and is similar to class-wise perturbation. Therefore, person-specific mask generation can increase efficiency while maintaining effectiveness. It is shown that EM, REM, and OPOM can support generating person-specific masks.
    
\item \emph{Effective for out-of-distribution data leakage}: In the real world, the attacker can not only access the user's protected images. For example, the attacker can extract the user's unprotected images from the photos shared by his friends. It will incur additional challenges for face privacy protection. Among the existing protection works, it is shown that only EM,  REM, APF, and Fawkes can also maintain the protection ability in such a case.

\end{itemize}
\begin{table*}[ht]
\centering
\caption{Comparison of the State-of-the-art Image Obfuscation-based Protection Methods.}
\label{tab:my-table1}
\setlength{\tabcolsep}{4.6mm}
\begin{tabular}{|c|c|c|c|c|c|c|c|c|}
\hline
Methods   &Whole image obfuscation &API &Open-set &\begin{tabular}[c]{@{}c@{}}Real-time\\capability \end{tabular}& Scalability &\begin{tabular}[c]{@{}c@{}}Hardware\\compatibility \end{tabular}\\ \hline
AMT-GAN \cite{hu2022protecting}     & NO      & YES           & YES  & NO               & Moderate            & NO  \\ \hline
Gender-AN \cite{tang2022gender}    & NO      & Not discussed & NO   & NO                    & Moderate            & NO  \\ \hline
DeepBlur  \cite{li2021deepblur}    &YES      & YES           & NO   & YES                    & High             & NO  \\ \hline

Diff-Privacy \cite{he2024diff}&YES&	YES	&YES&YES&High&NO\\ \hline
Optics \cite{lopez2024privacy}&YES&Not discussed &YES&YES&Low&YES\\\hline

Transferable \cite{zhangtransferable}&YES&YES&YES&YES&High&NO\\\hline

\end{tabular}
\end{table*}

\subsection{Performance Evaluation}
We compare the protection performance of two types of adversarial perturbation-based protection methods, as the codes for most image obfuscation-based methods are currently unavailable. 

We first evaluate EM, REM, and OPOM, all of which protect social media users' facial privacy without relying on auxiliary data.
The evaluation metrics are protection success rate (PSR) and similarity (SSIM). PSR is defined as the ratio of the number of wrongly recognized images to the total number of images, and SSIM refers to the structural similarity of the two images before and after protection. SSIM is similar to the human visual system and is sensitive to local structural changes. The larger the value, the more similar the protected image is to the original image for the human eye.

In the experiments, we utilized the CASIA-WebFace dataset, which comprises 494,414 images of 10,575 individuals. We randomly selected 50 identities to hide their identities and perturbed 80\% of the images for training the face recognition model, with the remaining 20\% of clean images used for testing. The perturbations were generated separately according to the original source codes. {\color{black}We set the uniform perturbation budget at 0.03.} The face recognition model structure is Inception$\_$Resnet, the training batch size is 64, and the epoch number is 50. The SSIM is calculated for all protected images before and after, with the average value obtained.

The comparison results of these four methods are shown in Fig.~\ref{fig:fl} and Fig.~\ref{fig:c}. 
It is shown that REM strikes a better balance between protection and image quality. The other two methods exhibit varying degrees of trade-off between protection and image quality. The low success rate of OPOM protection may stem from its limited number of images per identity. 
As shown in Fig.~\ref{fig:c} (left side), OPOM achieves favorable visual outcomes by using the $L_\infty$ norm to control the maximum value of perturbed elements. 
{\color{black}Regarding noise limitation, EM and REM exhibit comparable image quality.}

\begin{table*}[tt]
\centering
\footnotesize
\caption{Comparison of the State-of-the-art Adversarial Perturbation-based Protection Methods.}
\label{tab:my-table2}
\setlength{\tabcolsep}{2.7mm}
\begin{tabular}{|c|c|c|c|c|c|c|}
\hline
Methods &Auxiliary data requirement&API          & Open-set & Target-identity     & Person-specific & \begin{tabular}[c]{@{}c@{}}Effective for \\out-of-distribution data leakage\end{tabular} \\ \hline
EM  \cite{huang2021unlearnable}    & NO&Not discussed  & NO       & NO & YES             & YES           \\ \hline
REM  \cite{fu2021robust}   &NO& Not discussed  & NO       & NO & YES             & YES           \\ \hline
OPOM \cite{zhong2022opom}   & NO& YES           & YES      & NO           & YES             & Not discussed \\ \hline
DFPP \cite{shamshad2024makeup} &NO & Not discussed&YES&NO&YES& Not discussed\\ \hline

APF  \cite{zhang2020adversarial}   &YES & Not discussed  & YES      & NO           & NO              & YES           \\ \hline
Fawkes \cite{shan2020fawkes} &YES & YES           & NO       & YES           & NO              & YES           \\ \hline
TIP-IM \cite{yang2021towards} & YES &YES           & YES      & YES           & NO              & Not discussed \\ \hline
\end{tabular}
\vspace{-0.4cm}
\end{table*}

We then compare the protection performance of the following three auxiliary data requirement methods, i.e., APF \cite{zhang2020adversarial}, Fawkes \cite{shan2020fawkes}, and TIP-IM \cite{yang2021towards}. We also adopt protection success rate (PSR) and similarity (SSIM) as the evaluation metrics. 
SSIM is a Quality-of-Experience (QoE) metric that measures variations in brightness, contrast, and structure between the original and altered images.
{\color{black}According to the original papers, we utilized the LFW dataset, which includes 13,233 images collected from 5,749 individuals in natural scenes.} To ensure consistency, we standardized the image size to 112$\times$112$\times$3, set the learning rate to 0.01, the batch size to 1, and the epoch to 100. We then generated 500 protected images, each with a unique identity, through the use of three methods. Finally, we calculated the PSA by feeding the generated protected images to the ArcFace face recognition model. The Structural Similarity Index Measure (SSIM) was applied to 500 original and generated protected images, and the average value was derived. The comparison results are presented in Fig.~\ref{fig:2} and Fig.~\ref{fig:c} (right side).
It is shown that Fawkes can generate natural and artifact-free protected images, which can be attributed to its use of DSSIM for perceived similarity measurement. However, its protection success rate is low. APF has a high protection success rate but results in noticeable distortion patterns. TIP-IM strikes a balance between protection and image quality by incorporating the MMD (maximum average difference) loss.

%% file: future.tex
\section{Future Research Directions}

\subsection{Potential Research Directions on Image Obfuscation-based Protection Methods}

We propose the following research directions for enhancing the performance of image obfuscation-based protection methods.

First, it is shown that the facial images generated through image obfuscation sometimes can have a noticeable departure from the original image. Excessive modifications to facial features can have a negative effect on user experience. To improve the application, a more targeted approach that only changes the identity-relevant aspects of the face is needed. Further research should identify key facial characteristics impacting identity and modify those areas for greater privacy protection.


Second, makeup-based image perturbation methods \cite{hu2022protecting} have been primarily designed for female facial images. However, effective privacy protection should be universal. There is a need to expand these techniques to be equally effective for male facial images. Future research could explore gender-neutral perturbation techniques or develop makeup-based protection methods that work across different genders.

Third, image obfuscation will inevitably alter the appearance of facial images, thus causing visible artifacts. To effectively preserve privacy, face de-recognition methods should generate facial images that maintain a natural look and retain key facial features. Further research is needed to reduce the artifacts produced by these techniques.

Finally, existing facial privacy protection methods mainly focus on static photos. However, there is a research gap in protecting privacy in videos, which are dynamic and composed of multiple frames. How to design effective privacy protection methods for videos requires further exploration.

\subsection{Potential Research Directions on Adversarial Perturbation-based Protection Methods}

We propose the following avenues for exploration in the area of face privacy protection using adversarial perturbations.

Firstly, most current methods focus on selfies, which are clear and upright portraits. However, surveillance cameras capture people in varying light and occlusion conditions in reality. To better reflect real-life scenarios, future research should consider more challenging environments for face recognition. 
Additionally, it's crucial to consider privacy protection beyond static images, extending to dynamic video fields. Researchers may consider using advanced algorithms for face alignment or employing super-resolution methods to enhance key features such as the eyes.

Second, adversarial perturbation-based privacy protection methods often lead to image quality degradation. It is important to ensure that the protected image maintains a good visual experience. Future research should design protection tools that do not negatively impact the visual experience of users. For instance, instead of relying on traditional norms such as the $L_2$ and $L_\infty$ norm, researchers may consider alternative constraints such as reducing the number of modified pixels or even trying the one-pixel attack approach.

Third, to make the protection methods accessible to users, the time cost of generating protected images must be reasonable.
A potential research direction is to improve the noise generation process for higher efficiency. A potential solution is to restrict noise distribution to specific areas, such as the face's position and size, rather than adding noise to the entire image.

{\color{black}Lastly, in the future, new attack threats may emerge that can circumvent current privacy protection measures. While existing defensive perturbations have proven effective, in real-world training at commercial companies like Google and Alibaba, expert trainers often include pre-processing procedures (e.g., data augmentation) in collected samples to improve model performance. Data augmentation enhances both data quantity and adversarial robustness. To this end, these protected faces might still be identifiable to face recognition models after advanced augmentation. Moreover, adversarial training during the training process may also weaken existing protection. It is essential to address these potential challenges and develop more comprehensive and long-term protection plans.}

%% file: conclusion.tex
\section{Conclusion}

The leakage of facial information in cloud-based services poses significant security risks. In this article, we present a comprehensive overview of current methodologies for protecting facial image privacy. We categorize these methods into two primary approaches and provide an in-depth discussion of each. Additionally, we conduct both qualitative and quantitative evaluations of various protection techniques. Finally, we identify open challenges and propose potential future research directions to enhance privacy preservation in cloud environments.